\documentclass{article}

\usepackage{amsmath}
\usepackage{amssymb}
\usepackage{mathtools}
\usepackage{amsthm}
\usepackage{multirow}
\usepackage{soul}
\usepackage{xcolor}
\definecolor{myblue}{RGB}{204,229,255}

\usepackage[nice]{nicefrac}
\theoremstyle{plain}

\theoremstyle{definition}

\theoremstyle{remark}

\usepackage{url}
\usepackage{booktabs} 

\usepackage[accepted]{icml2023}

\usepackage{adjustbox}
\usepackage{multicol}
\usepackage{subfigure}

\icmltitlerunning{Learn What \textbf{NOT} to Learn: Towards Generative Safety in Chatbots}

\begin{document}

\twocolumn[
\icmltitle{Learn What \textbf{NOT} to Learn: Towards Generative Safety in Chatbots}



\icmlsetsymbol{equal}{*}

\begin{icmlauthorlist}
\icmlauthor{Leila Khalatbari}{yyy,Sharif}
\icmlauthor{Yejin Bang}{yyy}
\icmlauthor{Dan Su}{yyy}
\icmlauthor{Willy Chung}{yyy}
\icmlauthor{Saeed Ghadimi}{Waterloo}
\icmlauthor{Hossein Sameti}{Sharif}
\icmlauthor{Pascale Fung}{yyy}

\end{icmlauthorlist}

\icmlaffiliation{yyy}{Department of Electrical Engineering, Hong Kong University of Science and Technology}
\icmlaffiliation{Sharif}{Department of Computer Engineering, Sharif University of Science and Technology}
\icmlaffiliation{Waterloo}{Department of Management Sciences, University of Waterloo}

\icmlcorrespondingauthor{Leila Khalatbari}{lkhalatbari@connect.ust.hk}

\icmlkeywords{Machine Learning, ICML}

\vskip 0.3in]



\printAffiliationsAndNotice{}  

\begin{abstract}
Conversational models that are generative and open-domain are particularly susceptible to generating unsafe content since they are trained on web-based social data. Prior approaches to mitigating this issue have drawbacks, such as disrupting the flow of conversation, limited generalization to unseen toxic input contexts, and sacrificing the quality of the dialogue for the sake of safety. In this paper, we present a novel framework, named "LOT" (\textbf{L}earn n\textbf{OT} to), that employs a contrastive loss to enhance generalization by learning from both positive and negative training signals. Our approach differs from the standard contrastive learning framework in that it automatically obtains positive and negative signals from the safe and unsafe language distributions that have been learned beforehand. The LOT framework utilizes divergence to steer the generations away from the unsafe subspace and towards the safe subspace while sustaining the flow of conversation. Our approach is memory and time-efficient during decoding and effectively reduces toxicity while preserving engagingness and fluency. Empirical results indicate that LOT reduces toxicity by up to four-fold while achieving four to six-fold higher rates of engagingness and fluency compared to baseline models. Our findings are further corroborated by human evaluation.
\end{abstract}
\section{Introduction}
\begin{figure*}
\centering
\includegraphics[scale=0.5]{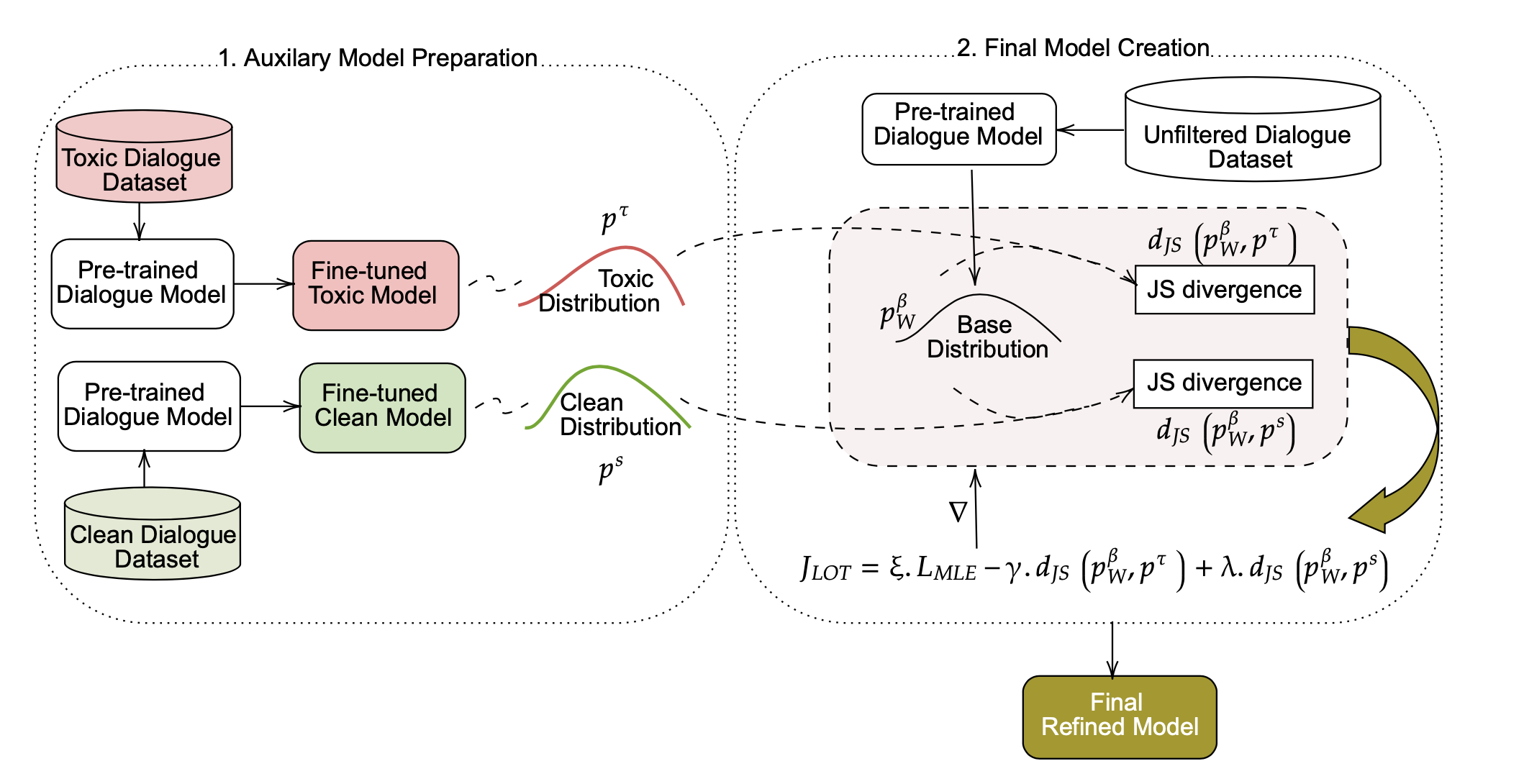}
\caption{Framework Overview of LOT}
\label{fig:overview}
\end{figure*}
The field of natural language processing (NLP) has witnessed significant advancements with the emergence of Large Language Models (LLMs), including GPT-2 \cite{gpt2} and GPT-3 \cite{gpt3}, PaLM \cite{palm}, XLNet \cite{xlnet}, and BERT \cite{bert}, as well as versatile dialogue models such as BlenderBot \cite{blenderbot3}, LaMDA \cite{lambda}, Dialogpt \cite{dialogpt}, and Sparrow \cite{glaese2022sparrow}. These models have experienced a paradigm shift in generative power through the incorporation of transformer architecture \cite{vaswani2017attention} and parameter scaling. Despite their impressive performance, these models are prone to learning undesirable behaviors such as toxicity, bias, and offense, as they rely on human-generated web data for training \cite{schramowski2022large,korbak2023pretraining,webDataToxicity, webDataToxicityb}. This predicament raises significant safety concerns, particularly when these models interact directly with humans. Therefore, the development of safe dialogue models has become an urgent research problem that requires immediate attention. However, the challenges associated with this endeavor are formidable. Existing approaches towards safe generation often necessitate the sacrifice of essential dialogue features such as relevance, diversity, engagement, and fluency.

Within the domain of toxicity mitigation, initial techniques employed n-gram blocking to sift through text and remove lexemes belonging to a predefined toxic vocabulary \cite{nGramBlocking1,ngramBlocking2}. However, these techniques were deemed insufficient for detecting subtler forms of toxicity, such as instances where an apparently innocuous sentence or phrase conveys a toxic meaning. To mitigate this issue, a group of methods emerged that prioritize the re-ranking of model outputs based on toxicity scores generated by a discriminative model \cite{dathathri2019plug}.  \cite{meade2023using} also rank multiple candidate responses based on a measure they call \textit{safety demonstrations}. One notable shortcoming of these approaches is that they may encounter difficulties producing relevant output for input text that falls below a specified toxicity threshold, especially in the presence of user-generated toxic content. Additionally, researchers have endeavored to improve the performance of generative models by cleansing input data and fine-tuning models using exclusively clean data. Nonetheless, a significant challenge associated with this approach is that models trained solely on clean data may not be able to generalize adequately when faced with toxic input during practical use.

The upsurge in the magnitude of model parameters stimulated a mounting interest in inference-time strategies to mitigate toxicity \cite{gedi, paraGedi, dexperts, marco}. Inference-time approaches do not necessitate any training or fine-tuning, but they do entail additional processing overhead during the decoding stage, which could trigger delays or require more memory resources. Furthermore, some approaches have been suggested to fine-tune a generative model using a contrastive or regularized loss \cite{cringe}. The benefit of these training-time strategies is that they empower the model to generate safe responses more frequently, thereby rendering it intrinsically safe. Consequently, it is plausible to employ an inference-time method or RLHF\footnote{Reinforcement Leaning with Human Feedback} on a well-tuned safe model for further toxicity alleviation. Nevertheless, if the language model has not been fine-tuned with a well-suited loss or data, it could impede the quality of generated responses in terms of relevance, diversity, fluency, and engagement.

This paper introduces a contrastive framework for dialogue generation, aimed at tackling the aforementioned issues and challenges. Our proposed methodology, named LOT (\textbf{L}earn What n\textbf{OT} to Learn), introduces additional training signals to guide the model towards generating safe responses by regulating the standard loss. The regularization terms utilize divergence measures to pull the model distribution far from the unsafe sub-spaces and closer to the safe regions in the hypothesis space. The safe and unsafe reference distributions are learned from positive (safe) and negative (toxic) samples, as illustrated in Figure 1. Our divergence terms consider the safe and toxic distributions in parallel for each input data entry.

We demonstrate the efficacy of our safety approach through automatic and human evaluations. LOT proves to reduce toxic generations while preserving (enhancing in some cases) the dialogue quality in terms of fluency, diversity, engagingness, and conversation flow. Furthermore, our method exhibits faster decoding times compared to baselines such as Defender \cite{defender} and PPLM \cite{dathathri2019plug} and requires less memory than other approaches such as GeDi \cite{gedi}, DEXPERTS \cite{dexperts}, and MarCo \cite{marco}. Additionally, our approach is also cost-efficient and intuitive, with a simple implementation that requires only a few finetuning epochs. LOT is agnostic of the type of language models being used either as the backbone model or the auxiliary models. Moreover, LOT is also task agnostic and can be easily applied to other tasks rather than the safety of dialogue generation. In summary, our contributions in this paper are fourfold:

\begin{itemize}
    \item We provide an extra training signal to the model using the automatically generated negative distribution parallel to each training data entry. 
    \item We provide an effective contrastive loss by adding divergent terms to the standard loss to guide the generations in the hypothesis space.
    \item We demonstrate that our controlled generation strategy can keep the conversation flow and consequently keep the user engaged. 
    \item We provide evidence that our framework decreases the toxicity of the generations while keeping or enhancing fluency, relevance, and engagingness.
\end{itemize}

We show that our model outperforms a set of recent baselines including Bot\_Adverserial\_Dialogue \cite{BAD}, Defender \cite{defender}, and different versions of fine-tuned BlenderBot.
\section{Related Work}\label{sec:related}
We divide the toxicity mitigation methods into two main categories of safe generation strategies and safe decoding strategies. Safe generation methods, directly output the safe response. In contrast, safe decoding approaches, generate regular responses and subsequently apply a safety mechanism during inference to convert any unsafe generations into safe responses.
\subsection{Safe Generation Strategies}
Safe generation is enforced through training or fine-tuning a model. The primary means of achieving safe generation during training is by using either a safe objective function or a designated dataset. 
\paragraph{Designated dataset:} A number of research efforts have been dedicated to designing datasets to incorporate baked-in safety into a model during fine-tuning \cite{recipe, ctrl}. Precisely, they engineer responses corresponding to toxic samples to teach the model how to react to users' toxic input. Some other solutions collect data in an adversarial manner and replace toxic responses with a safe canned (template) sentence. This data is further employed to fine-tune the model to boost safety \cite{BAD}. \cite{paradetox} has introduced a pipeline to gather parallel toxic-neutral sentence pairs via paraphrasing by crowdsourcers. This data is later used to train a model to generate less toxic and more neutral utterances.   

\paragraph {Safe objective functions:} The focus of these methods is to develop or regularise objective functions to ensure safer generations. This can be achieved by regularisation of the loss function. Additionally, contrastive learning paradigms introduce objectives that allow for learning from both positive and negative samples \cite{cringe, defender2}. The negative samples are conventionally synthesized and impact the quality of generations. \cite{DiscriminativeLatentSpace} projects the samples from the original latent space to a controllable discriminative-latent space by training a projection layer and a discriminator. The final discriminative-latent space can be well-separated by a target attribute (toxicity). 

\subsection{Safe Decoding Strategies}
Safe decoding strategies do not generate the safe response directly from the model. Rather, they manipulate the output distribution of the model at inference time to further transform it into a safer response. 
\paragraph{Detection-based methods:} While detecting toxicity is generally a more straightforward task than mitigating it, detection-based methods are limited to identification and are unable to take measures to neutralize the toxic content. Instead, detected toxic responses are typically replaced with a pre-written, off-topic template sentence. Although this approach effectively blocks the generation of unsafe utterances, it also undermines the primary objective of a chatbot: to simulate engaging and natural conversation. To address these limitations, researchers have recently proposed attack-defense mechanisms \cite{defender}. The attack is designed to maximize the likelihood of the toxic tokens with the aim of data collection. The defense aims to neutralize the attack once it is detected.
\paragraph{Decoding-time methods:} embed their toxicity mitigation strategy inside the decoding procedure to steer the original output towards a safer distribution. 

In accordance with decoding-time techniques, several methods have been developed to manipulate initial output logits in order to mitigate toxicity. DEXPERTS \cite{dexperts}, for instance, achieves this by utilizing the logits of an expert and anti-expert LM, both of which receive the same input utterance. In contrast, MARCO \cite{marco} leverages expert and anti-expert LMs to estimate the divergence between their respective token distributions. This enables the model to identify toxic tokens and subsequently regenerate them. Meanwhile, GeDi \cite{gedi} assigns weights to the output distribution at inference time based on the extent to which each token exhibits a certain attribute such as toxicity. These weights are calculated using Bayes rule over the output of an LM and a discriminator for the desired attribute. The probability of the discriminator is obtained by utilizing two CC-LMs\footnote{Class Conditional Language Models} with control-code and anti-control code, similar to the expert and anti-expert LMs in \cite{dexperts}.

ParaGeDi \cite{paraGedi} offers two solutions for text detoxification. Their first solution follows the GeDi formulation of Bayes rule over LM and CC-LMs but replaces the regular LM with a paraphraser. Their second solution utilizes conditional BERT. However, instead of a random selection of the words to be masked, they find the toxic words. A logistic regression model serves as the classifier to identify the toxicity of each sentence and its weights are the corresponding words. The words having the highest weights are considered toxic. 

ParaGeDi \cite{paraGedi}, presents two methods for text detoxification. The first approach adopts the GeDi formulation of Bayes rule over the main LM and CC-LMs. However, unlike GeDi, ParaGeDi replaces the regular LM with a paraphraser. In the second approach, a conditional BERT is employed to identify and mask toxic words instead of randomly masking words. For this purpose, a logistic regression classifier evaluates the toxicity of each sentence, with the weights assigned to the corresponding words. The words with the highest weights are deemed toxic.

The major limitation of decoding-time approaches is their slower decoding which makes them inappropriate for dialogue tasks. They are also memory-expensive as they need to keep the main model along with the auxiliary safety module in memory during the usage of the system. 

Decoding-time approaches are subject to constraints that render them unsuitable for dialogue tasks. Some of them suffer from slow decoding \cite{ctrl, dathathri2019plug} and some are memory-intensive as they require keeping the auxiliary safety module in memory as well as the main model \cite{dexperts, gedi, paraGedi, marco}. during system usage. As a result, these limitations significantly restrict the practicality of decoding-time approaches for use in dialogue tasks.
\section{Problem statement}
\label{sec:Ps}
In this section, we explain the preliminary concepts, provide the fundamental definitions, state the problem we aim to solve and elaborate on our framework. 
\subsection{preliminary concepts}\label{subsec:priliminary}
The end-to-end generative dialogue systems learn the latent mapping between the input and the corresponding output language employing a vast amount of open-domain unsupervised data \cite{lambda,blenderbot3,dialogpt}. The input to such models is a certain length of the conversation history and their output is the probability of the next token given the current history. The training of such dialogue systems is done through either auto-encoder masking or next token auto-regressive optimization paradigm in order to generate the best content in the context of an ongoing conversation. 
\vspace{-10pt}
\paragraph{Definition 1.}(Next token distribution)
The distribution of the next token is estimated via maximum likelihood as in equation \ref{eq:MLE}. 
\vspace{-10pt}
\begin{equation}\label{eq:MLE}
   L_{MLE}^{(i)}(p_\Theta,x^{(i)},y{(i)})=-\sum_{t=1}^{|y^{(i)}|}log p_\Theta(y_{t}^{(i)}|x^{(i)},y_{<t}^{(i)})
\end{equation}
Where  $\chi^D={(x^{(i)},y^{(i)})}$ is the dataset, $x^{(i)}$ is the input context and $y^{(i)}$ is the next target content. 
\vspace{-10pt}
\paragraph{Definition 2.}(Toxic, safe, and base Conversational Models) We denote a conversational model by its parameter set, $\theta^{j}$ where $j\in \{\tau, s, \beta, o\}$ determines if the model is toxic, safe, base, or output. The base model is a default pre-trained conversational model. The output model is the final detoxified conversational model attained from our framework. The toxic and safe models are those fine-tuned on toxic and clean data correspondingly. Each model of $\theta^{j}$, is associated with an output distribution of $p^{j}_{\theta}$ and a set of generations denoted by $G^{j}_{\theta}$.
\paragraph{Problem}(diminishing toxic generations) Given $\theta_{\beta}$ as the base conversational model, we aim to adjust $w\in\theta$ to minimize toxic generations in $G^{o}$, corresponding to the output model $\theta^{o}$ as the output model while maintaining the user-pertinent associations with regard to fluency, relevance, diversity, and engagingness. The following equations, mathematically define what we aim for by solving the defined problem.
\begin{flalign*} 
f_{tox}(G^o) << f_{tox}(G^\beta)\\ 
f_{ppl}(G^o)) \approx  f_{ppl}(G^\beta)\\ 
Max_{w\in \theta^o_W}(\tilde{d}(p^o_W(y|x),p^\tau(y|x)) \\ 
Min_{w\in \theta^o_W}(\tilde{d}(p^o_W(y|x),p^s(y|x))
\end{flalign*}

Here, $\tilde{d}(.)$ represents any divergence between its input arguments. $f(.)$ denotes any function that measures the quantity of its subscripts (perplexity or toxicity). Any model with the subscript $W$, participates in the optimization process, and its parameters change toward the optimum. The models without the subscript of $W$, are frozen and their parameters won't change in the course of optimization. 
\subsection{Framework overview}
Figure \ref{fig:overview} illustrates the overview of our proposed method dubbed LOT which aims to mitigate toxic generations of open-domain dialogue systems. In this context, it is highly desired to have a model which is created through a low-cost process, fast at decoding, and capable of generating more neutral replies when exposed to user's toxic contents and topics. We employ a new loss (called LOTLoss after the name of our method) composed of regularization terms to learn from both negative and positive (toxic and safe) samples in a contrastive manner to control the generation and steer the distribution towards safer regions. 

Precisely, we employ two auxiliary LMs, $\theta^s$ and $\theta^\tau$ to output parallel positive and negative distributions for each data entry, $x^(i)$. Then Jensen–Shannon (JS) divergence is utilized to guide the learning LM, $\theta^\beta$, towards the safe distribution, $p^s_\theta$, and away from the toxic regions, $p^\tau$, in the hypothesis space.

The obtained final distribution from this framework, $p^o$, is expected to generate less toxic content. 

In comparison to the decoding-time safety solutions, our approach does not create any extra time lag to achieve safer generation at decoding and thus is well-suited for dialogue applications. Moreover, LOT does not require any additional memory and GPU resources for the auxiliary safety models during inference.

\section{Methodology: Contrastive Learning Framework}
 To solve the problem elaborated in section \ref{subsec:priliminary}, we have proposed a contrastive loss function containing two regularization terms to fine-tune the base model, $\theta^\beta$. The role of the regularization terms is to pull away the distribution of the base model, $p^\beta(y|x)$, from the toxic data distribution, $p^\tau(y|x)$ and concurrently push it closer to the safe data distribution, $p^s(y|x)$. To pursue this, we have utilized the probabilistic distance function of Jenson-Shannon ($JS$) divergence \cite{js}. The contrastive framework is beneficial as it takes a large number of positive samples to teach a model what not to generate. However, the model can learn the undesired content with a few negative samples. As a result, incorporating negative samples provides an extra training signal over a fixed dataset size and improves the generalization of the model \cite{cringe}. 
 
Unlike the standard contrastive learning frameworks \cite{cringe, romain}, LOT does not require synthesis of parallel positive and negative samples that greatly affect the quality of learning. 
Instead, we automatically generate positive and negative data distributions parallel to the input data that is required to compute the regularization terms of LOT loss, $J_{LOT}$. Equation \ref{eq:CN_loss} describes the LOT proposed loss.
 
 \begin{equation}\label{eq:CN_loss}
    J_{LOT}=\xi.L_{MLE}-\gamma.d_{JS}(p^\beta_W,p^\tau)+\lambda .d_{JS}(p^\beta_W,p^s)
\end{equation}
Where the first term, $L_{MLE}$ is the standard CE loss of equation \ref{eq:MLE} that guarantees the next token to be the most probable according to the learned distribution, syntax, and semantics of the corresponding language. The terms $d_{JS}$ in $J_{LOT}$ of equation \ref{eq:CN_loss}, compute the JS divergence between the distribution of the learning model and a source distribution (toxic distribution in the second term and safe distribution in the third term). Having $d_{JS}$ terms in $J_{LOT}$ during optimization regulates the model weights so that the next token distribution is closer to safe regions and further from the toxic regions while being sensical according to the conversation history. 

$JS$ divergence measures to what extent two or more distributions can be determined from one another. $JS$ divergence is composed of two Kullback-Liebr (KL) divergence terms \cite{kl} while having two advantages over KL. First, $JS$ divergence has always finite values even for infinite random variables. Second, it is a symmetric distance contrary to $KL$ divergence. These properties make $JS$ a better fit for our optimization process as infinite divergence values can obstacle backpropagation and the asymmetry can lead to inconsistent formulation if the order of arguments is interchanged. Considering two distributions of $q$ and $p$, $JS$ divergence is defined as follows:
\begin{equation}\label{eq:JS}
\begin{split}
    d_{JS}(p^\beta \parallel p^i)=\frac{1}{2}d_{KL}(p^\beta \parallel \Lambda )+\frac{1}{2}d_{KL}(p^i\parallel \Lambda)\\
\Lambda =\frac{1}{2}(p^\beta+p^i),  i\in\{\tau,s\}
\end{split}
\end{equation}
$JS$ divergence values fall in $[0,1]$ if one of the distributions is of base 2 logarithm, which is considered a form of normalization for this metric.

$KL$ divergence is also a probabilistic distance measuring how different a given distribution is from a reference distribution. Interpretation of $KL$ between two distributions of $p$ and $q$ is the expected excess surprise from using $q$ as a model when the actual distribution is $p$. $KL$ divergence is a distance but not a metric as it does not satisfy symmetry and triangle inequality conditions. 
\begin{equation}\label{eq:HL}
\begin{split}
    d_{KL}(p^\beta \parallel p^i)=\sum_{x\in X}p^\beta(x)log\frac{p^\beta(x)}{p^i(x)}=\\-\sum_{x\in X}p^\beta(x)log\frac{p^i(x)}{p^\beta(x)}  , i\in\{\tau,s\}
\end{split}
\end{equation}
It is the expectation of the logarithmic difference between the probabilities $p^\beta$ and $p^i$, where the expectation is taken using the probabilities $p^\beta$.
 $KL$ divergence is defined so only if for all $x$,  $P^i(x)=0$ implies $p^\beta(x)=0$ (absolute continuity). Whenever $p^\beta(x)$ is zero the contribution of the corresponding term is interpreted as zero because  $\lim_{x\rightarrow 0^+}xlog(x)=0$. The domain of $KL$ divergence values fall in $[0,\infty]$. 

Each term in the proposed loss of equation \ref{eq:CN_loss}, is multiplied by a coefficient that is intended to control two factors. First, balancing the value range of each term in the loss so that the scale difference does not intensify the impact of one term and attenuates the impact of the others during the optimization process. Second, the coefficients determine how much we want to sacrifice coherence and relevance to safety. 
\begin{table}
\centering
\begin{tabular}{lccc}
\toprule Category & Train & Valid & Test \\ \midrule
        Safe Utterances & 42049 & 4239 & 1654 \\
        Offensive Utterances & 27225 & 2763 & 944 \\\midrule
        Total Utterances & 69274 & 7002 & 2598 \\
        Total Conversations & 5080 & 513 & 191 \\
\bottomrule
\end{tabular}
\caption{\label{table:BAD} BAD dataset statistics }
\end{table}
\begin{table*}[]
\centering
    \begin{adjustbox}{width=0.7\linewidth,totalheight={\textheight},keepaspectratio}
    \begin{tabular}{ccccc}
    \toprule
    \multicolumn{1}{c}{\textbf{Model}} & \multicolumn{1}{c}{\textbf{Fluency}} & \multicolumn{1}{c}{\textbf{Toxicity} } & \multicolumn{2}{c}{\textbf{Diversity}} \\ \cmidrule(l{2pt}r{2pt}){2-2} \cmidrule(l{2pt}r{2pt}){3-3} \cmidrule(l{2pt}r{2pt}){4-5} 
     & \multicolumn{1}{c}{\textbf{PPL$\downarrow$}} & \multicolumn{1}{c}{\textbf{ParlAI Tox$\downarrow$}}  & \multicolumn{1}{c}{\textbf{Canned. Sent (\%)$\downarrow$}} & \multicolumn{1}{c}{\textbf{Div1$\uparrow$}}\\ \midrule
     LOT$_{re/con/JS}$ & \multicolumn{1}{c}{7.312} & \multicolumn{1}{c}{\textbf{0.610}} & \multicolumn{1}{c}{0.000} & \multicolumn{1}{c}{17.403$^{+}_{-}2.320$} \\ 
    LOT$_{contraster}$  & 7.712 & 1.270 & 0.000 & $17.206^{+}_{-}3.180$\\
    LOT$_{reinforcer}$ & \textbf{6.976} & 1.229 & 0.000 & \textbf{17.655$^{+}_{-}2.510$}\\
    LOT$_{re/con/KL}$ & 7.131 & 0.656 & 0.000 & 17.467$^{+}_{-}2.320$\\\bottomrule
    \end{tabular}
    \end{adjustbox}
    \caption{Ablation study on LOT. LOT$_{re/con/JS}$ denotes the final LOT that utilizes \textit{JS} as its divergence, and its loss is comprised of three contrastive components. LOT$_{contrastor}$, represents LOT employing \textit{JS} divergence, but only two components of \textit{CE}, and contrastor in its loss. LOT$_{reinforcer}$ indicates LOT with \textit{JS} and two components of loss components including \textit{CE} and reinforcer. LOT$_{re/con/KL}$ denotes LOT with three-component loss and \textit{KL} as its divergence.} 
    \label{tab:ablation}
\end{table*}

\begin{table*}[]
\centering
    \begin{adjustbox}{width=0.7\linewidth,totalheight={\textheight},keepaspectratio}
    \begin{tabular}{lcccc}
    \toprule
    \multicolumn{1}{c}{\textbf{Model}} & \multicolumn{1}{c}{\textbf{Fluency}} & \multicolumn{1}{c}{\textbf{Toxicity} } & \multicolumn{2}{c}{\textbf{Diversity}} \\ \cmidrule(l{2pt}r{2pt}){2-2} \cmidrule(l{2pt}r{2pt}){3-3} \cmidrule(l{2pt}r{2pt}){4-5} 
     & \multicolumn{1}{c}{\textbf{PPL$\downarrow$}} & \multicolumn{1}{c}{\textbf{ParlAI Tox$\downarrow$}}  & \multicolumn{1}{c}{\textbf{Canned. Sent (\%)$\downarrow$}} & \multicolumn{1}{c}{\textbf{Div1$\uparrow$}}\\ \midrule
    BB\_BAD & 15.671 & 2.825 & 0.730 &  18.650\\
    BB\_BAD\_clean & 7.963 & 1.163   & 0.150 &  18.015\\
    BB\_vanilla & 14.766 & 3.130  & 10.170 &  \textbf{23.333}\\ \midrule
    Defender & 8.113 & 2.480  & 0.008 &  12.288\\ \midrule
    LOT (Ours) & \multicolumn{1}{c}{\textbf{7.312}} & \multicolumn{1}{c}{\textbf{0.610}} & \multicolumn{1}{c}{\textbf{0.000}} & \multicolumn{1}{c}{17.403}\\  \bottomrule
    \end{tabular}
    \end{adjustbox}
    \caption{Effectiveness of LOT compared against the baselines in terms of perplexity, toxicity, and diversity}
    \label{tab:main_results}
\end{table*}
\section{Experiments}
\label{sec:experiments}
\subsection{Experimental setup}
\label{subSec:Experimental_setup}
Our target task in this paper is to mitigate unsafe content generation by dialogue models in the single-turn conversation scenario. 
We conducted a series of experiments on a dialogue dataset to assess the effectiveness of our proposed approach in reducing the incidence of toxic outputs while utilizing minimal computational resources. Our backbone dialogue model is BlenderBot-small provided by huggingface. 
The experiments were conducted on a server equipped With 2.10GHz Intel(R) Xeon(R) E5-2620 core i7 CPU, 125 GiB of RAM and GPUs of NVIDIA GeForce GTX 1080 with 11178 MiB memory.  
\paragraph{Dataset} To investigate our framework, we utilized the Bot Adversarial Dataset of dialogues, dubbed BAD \cite{recipe}. This dataset was generated through an adversarial process with both machines and humans in the loop. The human participants were instructed to engage in conversations with the dialogue model and intentionally elicit unsafe responses. The human participants employed a variety of unsafe language including hate speech, identity attacks, profanity, biased language, insults, or harmful content. Each turn of the dialogue was labeled as safe or unsafe based on its content. Table \ref{table:BAD} elucidates the statistics of the BAD.
\paragraph{Evaluation Metrics}include 4 dimensions of human judgments and 4 automatic metrics to give a comprehensive perspective of the potentials and breakpoints of our model compared to the introduced baselines.

\textbf{\- Automatic evaluation:}
\begin{itemize}
\item \textbf{Fluency:} We indicate the fluency of our model by measuring the perplexity of its generations. The value of perplexity quantifies the degree of resemblance between the learned distribution and the natural language distribution.
\item \textbf{Toxicity:} We measure the toxicity level of our model's generations by Parlai toxicity classification which is recognized for its heightened sensitivity to unsafe or inappropriate content \cite{parlai}. 
\item \textbf{Diversity:} 
We measure diversity in two distinct levels: the average sentence level diversity and the across-dataset level diversity. Sentence level diversity counts the unique number of n-gram occurrences (we use uni\-grams here) within a given dialogue turn \cite{diversity}. We define the cross-dataset diversity to measure how many canned (template) sentences are generated by a model across the entire dataset. 
\end{itemize}
\textbf{Human evaluation}
The automatic measures that are commonly employed to assess the performance of safe dialogue generation do not always align with human judgments about the conversation quality. Therefore, it is crucial to supplement with human evaluation scores as well. We further evaluate our model by AMT\footnote{Amazon Mechanical Turk} human annotators across four aspects of the generated response: (1)~\textbf{fluency} measures whether the response is coherent, syntactic, and semantic; (2)~\textbf{topical} measures if the response is a relevant and likely continuation to the context, (3)~\textbf{toxicity} measures if the response contains profanity, threat, hate speech, violence, insult, identity attack, [professional, medical, or harmful] advice, [gender, racial, religious] bias; (4) ~\textbf{engagingness} measures whether the generated responses can engage and keep the user in a long conversation. 
\paragraph{Baselines}
\begin{itemize}
    \item Defender \cite{defender}: presents a conversational framework consisting of five turns for both attack and defense. Their defense mechanism searches for tokens within the context and response that may have led to the unsafe generation. These identified tokens are subsequently masked and the response is regenerated by consuming the masked context. The reporting results corresponding to this baseline come from our own implementation from the original to accommodate scenarios involving single-turn conversations and defense-only contexts. 
    \item BB\_BAD: is referred to BlenderBot which we further fine-tuned on BAD dataset to embed a baked-in safety. Our motivation for including this baseline is to elaborate that the enhancement LOT offers is not simply due to another round of fine-tuning. Rather, the proposed contrastive loss of LOT is the source of improvement.
    \item BB\_BAD\_clean: is referred to BlenderBot that we fine-tuned merely on clean (safe) samples of BAD dataset. This serves as our baseline to substantiate and illustrate that the augmentation of clean data through fine-tuning can indeed enhance the safety of a model, albeit within certain constraints. However, we contend that LOT can confer even greater levels of safety to the model in comparison to the simple fine-tuning of the model with clean data.
    \item BlenderBot original: is the BlenderBot-small from \cite{BAD} and Hugging Face library. 
\end{itemize}
\subsection{Experimental results and discussion}
Table \ref{tab:ablation} presents the ablation study over different versions of LOT. We have compared how LOT works in the absence of the enforcer or contrastor. We also demonstrate how effective $JS$ divergence is compared to $KL$ divergence. The results indicate that the loss function composed of reinforcer, contrastor, and $JS$ divergence, denoted as $LOT_{re/con/JS}$, achieved the lowest toxicity while maintaining fluency and average utterance diversity (Div1). This verifies that each term of the proposed contrastive loss effectively contributes to the safety of the model. All versions of LOT, generate no canned sentences and thus have equal performance in terms of across-dataset diversity. The rest of the experiments are conducted with the best version  of LOT, $LOT_{re/con/JS}$.      

\begin{figure}[h]
\centering
\includegraphics[scale=0.4]{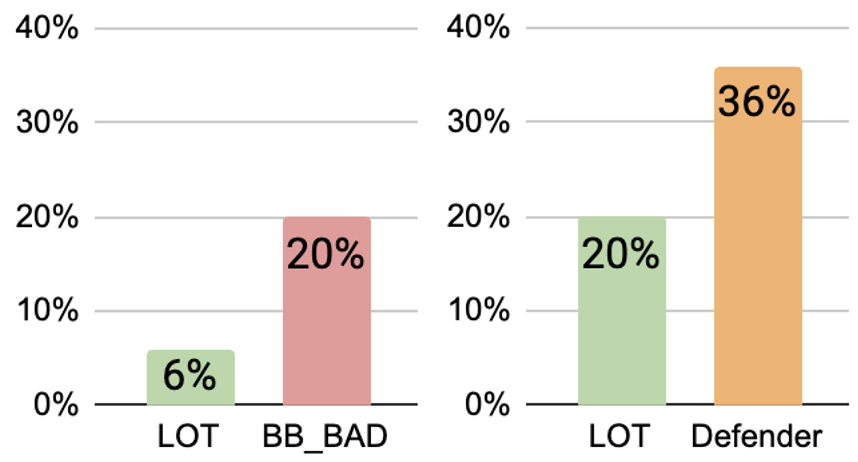}
\caption{ParlAi toxicity rates of LOT against two baselines of BB\_BAD and Defender. The results come from the A-B testing corresponding to the human evaluations}
\label{fig:toxic}
\end{figure}

\begin{figure}
    \centering
    \subfigure[LOT vs BB\_BAD]{\includegraphics[width=0.35\textwidth]{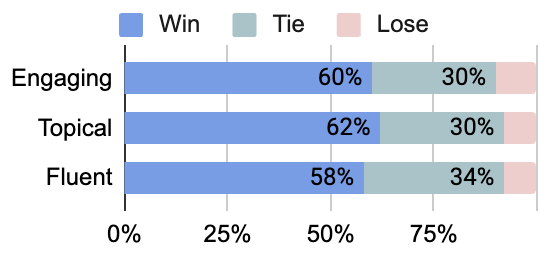}} 
    \subfigure[LOT vs Defender]{\includegraphics[width=0.35\textwidth]{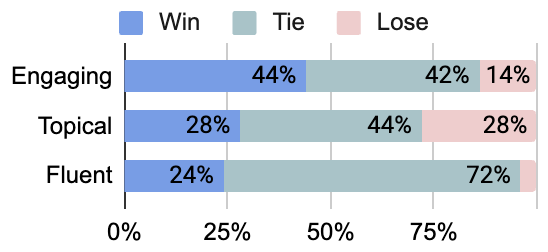}} 
    \caption{win lose, and tie rates of LOT compared against baselines corresponding to A-B testing results from Human evaluation}
    \label{fig:human_eval}
\end{figure}

Table \ref{tab:main_results} displays the effectiveness of LOT against the introduced baselines. 
\paragraph{Toxicity} Table \ref{tab:main_results} shows that LOT is the most effective method to reduce toxicity and improves safety by at least four times compared to Defender, BB\_vanilla, and BB\_BAD. Furthermore, we devised an experimental setup to assess the efficacy of fine\-tuning a model with clean data versus LOTLoss. To this end, we conducted a comparative analysis of BB\_BAD\_clean and LOT under identical conditions. The sole difference was that BB\_BAD\_clean was fine-tuned using clean (safe) samples from the BAD dataset, while LOT was fine\-tuned on the BAD dataset by LOTLoss. The results demonstrate that the toxicity of BB\_BAD\_clean was twice as high compared to LOT. Data cleaning and engineering require humans in the loop, is time-consuming, and cost-bearing. However, fine-tuning merely by clean data has a constrained potential to improve safety. This underscores the imperative of developing more efficient and effective frameworks, such as LOT. To further investigate the detoxification ability of LOT, we also conducted human evaluation over LOT versus BB\_BAD and LOT versus Defender. Figure \ref{fig:toxic} depicts the A\-B testing results comparing LOT with the two best-performing baselines. As per the evaluations by human assessors, LOT was deemed more than three times safer than BB\_BAD and nearly twice as safe as Defender.
\paragraph{Fluency:}As mentioned earlier, we reported perplexity to ensure that the generated text is syntactically correct, semantically meaningful, and conforms to the natural language distribution. In this paper, our primary objective was to reduce toxicity while maintaining fluency and other language-related attributes. Nevertheless, LOT not only enhances the safety of the model but also improves its fluency. Specifically, as presented in Table \ref{tab:main_results}, LOT is the most fluent and least toxic among the baselines. The human evaluation further corroborates the superior fluency of LOT. As per Figure \ref{fig:human_eval}, during A-B testing, LOT was found to be almost five times more fluent than BB\_BAD and nearly six times more fluent than Defender.
\paragraph{Diversity:} Canned\footnote{template} sentences that are replaced by detected toxic responses in some safety strategies are entirely off-topic and try to steer the conversation away from the unsafe ground in a very non-human-like manner. Frequent generation of canned answers destructs conversation-level diversity and makes the user lose interest to continue the conversation. This can be observed from the engagingness metric in Figure \ref{fig:human_eval}. 
To quantify the cross-dataset diversity, we introduced the \textit{Canned. Sent (\%)} metric in Table \ref{tab:main_results}. This metric represents the number of predetermined generated canned sentences by a given model, and it is inversely proportional to the degree of cross-dataset diversity. Consequently, a lower value of generated canned sentences indicates a higher level of cross-dataset diversity for the model. We observe that the decrease in the quantity of \textit{Canned. Sent (\%)} along different methods of table \ref{tab:main_results} is proportional to the reduction in uni-gram diversity, Div1. This suggests that part of Div1 sentence diversity comes from the canned sentences which are not desirable as they are entirely off-topic and disrupts the conversation flow.
\paragraph{Topical (relevance):} The degree of user engagement in a conversation is negatively impacted when the generated responses are off-topic and fail to relate appropriately to the input context. LOT exhibits comparable performance to Defender in generating topical sentences while significantly outperforming the BB\_BAD with a win rate of 60\%. This suggests that LOT can maintain or even enhance the relevance of generated content while minimizing toxicity.
\vspace{-6pt}
\paragraph{Overall:} In summary, the LOT dialogue framework not only reduces toxicity but also enhances the fluency and relevance of generated content. LOT outperforms the baselines in terms of fluency and has a higher level of cross\-dataset diversity due to its ability to avoid predetermined canned responses, which often disrupt the conversation flow. Moreover, LOT generates topical responses that are more engaging to the user compared to BB\_BAD and comparable to Defender. These results are corroborated by the human evaluation conducted in the study. Thus, LOT provides a comprehensive solution for generating safe, fluent, and relevant content in dialogues and meets the objectives of $f_{tox}(G^F) << f_{tox}(G^\beta)$ and  
$f_{ppl}(G^F) \approx  f_{ppl}(G^\beta)$, we defined in section \ref{subsec:priliminary}. 

Moreover, in contrast to decoding-time baselines such as Defender \cite{defender} and PPLM \cite{dathathri2019plug}, our model LOT exhibits a higher efficiency in terms of faster inference during decoding. Lot is also more memory-efficient during inference compared to methods such as \cite{dexperts}, Gedi \cite{gedi}, and \cite{marco} since these baselines keep extra auxiliary models in memory during decoding. The time and memory efficiency of LOT is attributed to the fact that we do not perform any safety operations during decoding, making LOT more suitable for the online task of dialogue.
\section{Conclusion and Future Direction}
The parameter scale of chatbots is strikingly growing. This parameter growth has led to an ever-increasing concern about the safety of the dialogue models. Despite the significance of toxicity mitigation in dialogue models, little effort has been  dedicated to dialogue-specific applications. The existing safe generation strategies suffer from sacrificing the generation's quality for safety. In this paper, we proposed LOT, a method for the safe generation of dialogue. LOT fine-tunes a backbone model with a contrastive loss that incorporates regularization terms. The two added regularisation terms try to minimize the $JS$ divergence to move closer to safer regions and stay away from toxic sub-spaces in the hypothesis space. In fact, LOT learns from both positive and negative distributions through the regularization terms. The experimental results and analysis show that LOT produces more fluent and engaging responses with higher cross-dataset diversity and fewer off-topic responses compared to the baselines while effectively improving the safety of the model. 

Our proposed framework, LOT, represents a significant step toward ensuring safety in open-domain dialogue. In future work, we plan to improve the LOTLoss function for better convergence while maintaining efficacy. We also aim to enhance the average n-gram diversity at the sentence level by modifying the current LOTLoss. Additionally, we will try to distill the safety capacity of LOT into a safety layer with fewer parameters to improve efficiency. Furthermore, we would like to expand LOT to a multi-turn conversation setting.



\newpage
\appendix
\onecolumn

\end{document}